\documentclass{article}

\PassOptionsToPackage{numbers,compress}{natbib}

\usepackage[dblblindworkshop, final]{neurips_2026}

\makeatletter
\renewcommand{\@noticestring}{Accepted at the NeurIPS 2026 AI for Drug Discovery workshop.}
\makeatother

\usepackage[utf8]{inputenc}
\usepackage[T1]{fontenc}
\usepackage{hyperref}
\usepackage{url}
\usepackage{booktabs}
\usepackage{amsfonts}
\usepackage{amsmath}
\usepackage{nicefrac}
\usepackage{microtype}
\usepackage{xcolor}
\usepackage{graphicx}
\usepackage{algorithm}
\usepackage{algpseudocode}
\usepackage{threeparttable}
\usepackage{wrapfig}

\title{\textbf{Inference-Time Steering Across Scaffolds and Epitopes for Sample-Efficient De Novo Nanobody Design}}

\workshoptitle{AI for Drug Discovery: Bridging the Translation Gap (NeurIPS 2026)}

\author{%
   Mason Minot, Simon Friedensohn, Cédric R. Weber \\
   Alloy Therapeutics\\
   Allschwil, Switzerland  \\
   \texttt{mason.minot@alloytx.com}
}

\begin{document}

\maketitle

\begin{abstract}
Generative molecular design typically relies on over-generation and post-hoc filtering. Inference-time steering methods like Feynman-Kac (FK) steering improve efficiency by biasing trajectories toward high-reward candidates, yet they optimize within a single fixed configuration. In \textit{de novo} nanobody design, upfront commitment to conditioning choices like scaffold identity or target epitope creates a bottleneck. Since yield varies across these conditions, steering unpromising setups wastes compute. We recast generative condition selection as an inference-time resource-allocation problem. We propose two dynamic budget allocators: a joint Global Sequential Monte Carlo (SMC) formulation that steers and allocates compute across configurations concurrently, and a decoupled approach adapting Batched Successive Elimination (BaSE). Empirically, these dynamic allocators increase the yield of designable candidates per GPU-hour nearly 5$\times$ without sacrificing diversity across distinct therapeutic targets and generative configurations.
\end{abstract}

\section{Introduction}
Diffusion and flow matching have made \textit{in silico} molecular design accessible \citep{stark2025boltzgen,rectorbrooks2026generalmultimodalproteindesign,didi2026scaling}. However, current methods rely on over-generation followed by computationally expensive post-hoc scoring and filtering pipelines (e.g., evaluating binding energy or interface confidence). Because the yield of viable candidates is often highly variable across different targets and configurations, campaigns expend a large portion of their computational budget evaluating true negatives, creating a bottleneck.

To reduce this waste, inference-time steering methods like Feynman-Kac (FK) steering \citep{singhal2025fksteering,skreta2025feynman} integrate scoring functions directly into the sampler without requiring model fine-tuning. By estimating denoised backbones via Tweedie's formula \citep{efron2011tweedie} and evaluating them against intermediate reward functions, FK algorithms sample from a reward-tilted distribution by actively cloning promising trajectories and culling low-scoring ones. While recent works apply FK steering to protein design \citep{rectorbrooks2026generalmultimodalproteindesign,hartman2025fkprotein,yoo2026searchbinder}, they optimize within a single generative configuration and leave specialized, scaffold-driven formats largely unexplored.

This poses a challenge for modalities like nanobodies (VHHs). Generative methods typically design sequences and conformations for the antigen-binding loops, known as complementarity-determining regions (CDRs), onto a fixed, user-specified framework \citep{bennett2026atomically, millefragoso2026germinal}. However, framework identity dictates paratope geometry, stability, immunogenicity, and generative yield \citep{vincke2009general, rossotti2022immunogenicity}. The scaffold must therefore be treated as a variable rather than a constant.

Consequently, trajectory success remains bottlenecked by upfront choices of the scaffold, target epitope, or hotspot conditioning, whose viability is unknown \textit{a priori}. This presents a budget-allocation problem \citep{karnin2013successivehalving,li2018hyperband,gao2019batched}. Steering exacerbates this issue, as relocating costly oracle evaluations into the generation loop means that committing to an unproductive configuration wastes budget and leaves viable conditions under-sampled.


\begin{figure}[!ht]
  \centering
  \includegraphics[width=0.68\textwidth]{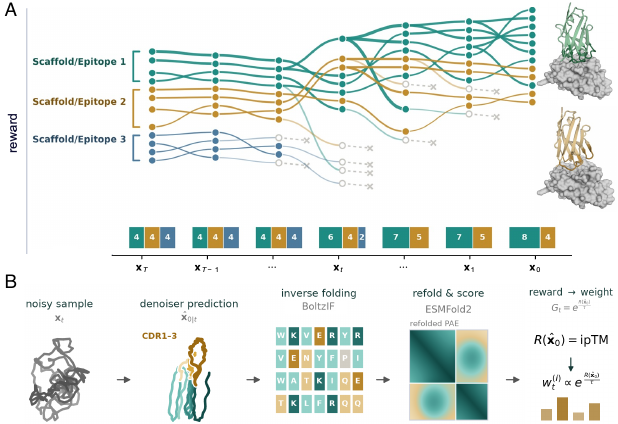}
\caption{\textbf{Overview of the inference-time steering and allocation pipeline.} \textbf{(A)} Illustration of the resource allocation and steering trajectory progressing from the initial noisy state $x_T$ to the final generated structure $x_0$. \textbf{(B)} Breakdown of the intermediate resampling step. At timestep $t$, Tweedie's formula is used to predict the denoised backbone $\hat{x}_{0|t}$ from the noisy sample $x_t$. The inverse folding model BoltzIF generates sequences for the designed CDRs. These are refolded with ESMFold2; the ipTM is then computed and used to update particle weights, $w_t^{(i)} \propto \exp\mathrm{(ipTM/}\tau)$.}
  \label{fig:steering_pipeline}
\end{figure}

\section{Methods}

\subsection{Fixed-Scaffold Nanobody Generation}
We formalize nanobody design as a conditional generation task, employing BoltzGen~\citep{stark2025boltzgen} to sample 3D CDR coordinates conditioned on a fixed scaffold. While standard generation samples candidates uniformly from the learned distribution $q_\theta(x_0)$, our goal is to shift this distribution toward high-quality, designable binders. Formally, we aim to approximate a reward-tilted target distribution over the full generative trajectory from initial noise at time $T$ to the final structure at time $0$:
\begin{equation}
    p_{\text{target}}(x_{0:T}) \propto q_\theta(x_{0:T})\exp(\Phi(x_{0:T})/\tau)
\end{equation}
where $q_\theta(x_{0:T})$ is the joint distribution of the unsteered generative trajectory, $\Phi(x_{0:T})$ is the reward path functional, and $\tau$ is a temperature parameter controlling the reward strength.
Candidate backbones generated from this distribution are evaluated for post-hoc designability by sampling 8 sequences per backbone using ProteinMPNN \citep{dauparas2022robust}, refolding each complex with AlphaFold2-Multimer (AF2-M) \citep{evans2021protein}, and scoring against AlphaProteo designability criteria \citep{zambaldi2024alphaproteo} (Appendix \ref{appendix:designability}). We employ an in-loop oracle (discussed below) that is orthogonal to our validation pipeline to prevent generation from over-optimizing for a single evaluator's biases.

\subsection{Feynman-Kac Steering via In-Loop ipTM}
Because exact sampling from $p_{\text{target}}$ is intractable, we apply Feynman-Kac (FK) steering over a parallel population of $N$ interacting particles. We interleave standard denoising steps with guidance events that evaluate particle quality, discarding unpromising states and replicating successful ones.

\begin{figure}[!ht]
  \centering
  \includegraphics[width=\textwidth]{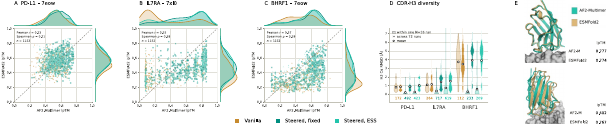}
  \caption{\textbf{Impact of inference-time steering on terminal quality and structural diversity.} \textbf{(A, B, C)} Oracle correlation (terminal ESMFold2 vs. terminal AF2-M) across three target-scaffold pairs, with marginal showing steering shifts toward higher ipTM. \textbf{(D)} CDR-H3 structural diversity of passing designs ($\text{C}\alpha$ RMSD); while steering narrows diversity within a run, it explores distinct structural basins across independent runs. Numbers denote total passing designs. \textbf{(E)} Structural examples of PD-L1 (7eow) candidates showing oracle agreement (top) and divergence (bottom).}
  \label{fig:fig2}
\end{figure}

At guidance event $m$ (diffusion timestep $t_m$), each particle's denoised terminal estimate $\hat{x}_{0, m} = f_\theta(x_{t_m}, t_m)$ is scored by an immediate in-loop reward $R(\hat{x}_{0, m})$. This scalar value is obtained by generating CDR sequences via BoltzIF, refolding the complex with ESMFold2~\citep{candido2026language}, and computing the interface predicted Template Modeling score (ipTM) (Figure \ref{fig:steering_pipeline}B). For independent steering baselines and the Successive Halving (SH) allocator, we adopt a memoryless \texttt{immediate} potential, assigning particles instantaneous normalized weights $w^i_m$ scaled by temperature $\tau$ and tracking the Effective Sample Size ($\mathrm{ESS}_m$) to monitor population collapse:
\begin{equation}
    w^i_m = \frac{\exp(R(\hat{x}_{0,m}^i)/\tau)}{\sum_{j=1}^N \exp(R(\hat{x}_{0,m}^j)/\tau)}, \qquad \mathrm{ESS}_m = \frac{1}{\sum_{i=1}^N (w^i_m)^2}
\end{equation}

We evaluate two resampling triggers: a fixed-cadence policy occurring at predetermined intervals tied to network time, and an adaptive policy dynamically triggered by weight degeneracy ($\mathrm{ESS}_m \le 0.5N$). Implementation details are provided in Appendix \ref{appendix:steering_background}.

\subsection{Condition Allocation Algorithms}

Steering individual trajectories fails to address the condition selection bottleneck. Because generative yield varies unpredictably across scaffolds and epitopes, standard baselines often waste compute on unpromising setups. To solve this, we introduce two dynamic allocation paradigms, both initialized with a uniform prior that distributes the starting computational budget equally across all candidate configurations:
    
\begin{itemize}
    \item \textbf{Joint Steering and Allocation via Global SMC:} Rather than steering $K$ isolated populations for $K$ candidate configurations (scaffolds or epitopes), we introduce a Global Sequential Monte Carlo (SMC) allocator that maintains a single, joint system of $N$ particles (Algorithm \ref{alg:mixture}). As ipTM values are bounded in $[0,1]$ and scaffold lengths are comparable, the allocator evaluates raw in-loop ipTM. We leave condition-specific score re-calibration to future work. Unlike the memoryless methods, this approach targets a cumulative reward path. At each shared guidance event $m$, the immediate reward increment $g^i_m = R(\hat{x}_{0,m}^i)$ is added to particle $i$'s historical log-weight and normalized globally across the entire population:
    \begin{equation}
    \label{eq:logw_update}
    \log w^i_m = \log w^i_{m-1} + \frac{g^i_m}{\tau} - \log \sum_{j=1}^{N} \exp \left( \log w^j_{m-1} + \frac{g^j_m}{\tau} \right)
    \end{equation}
    Consequently, no explicit budget is assigned. During global systematic resampling over the accumulated weights $\{w^i_m\}$, the computational budget emerges dynamically as unpromising particles are extinguished and successful ones replicate across condition boundaries.
    
    \item \textbf{Decoupled Allocation via Successive Halving (SH):} We formulate condition selection as a multi-armed bandit problem using the Batched Successive Elimination (BaSE) framework (Algorithm \ref{alg:bracket})\citep{karnin2013successivehalving,gao2019batched}. Configurations (arms) are steered independently, with the total computational budget allocated across successive rounds. We modify standard BaSE (detailed in Appendix \ref{app:bracket}) by: (1) distributing per-round budgets evenly among surviving arms, (2) ranking arms by the 90th percentile of their terminal validation scores (AF2-M ipTM), and (3) enforcing a deterministic 50\% elimination rate at the end of each round.
\end{itemize}

To isolate the effects of dynamic allocation from inference-time steering, we additionally evaluate ablation variants, specifically an unsteered SH approach and an SMC algorithm utilizing a fixed resampling cadence. Additional information and ablation results are detailed in the Appendix.

\begin{figure}[!ht]
  \centering
  \includegraphics[width=\textwidth]{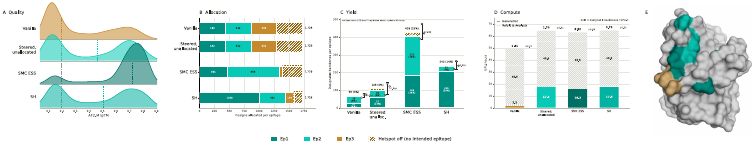}
  \caption{\textbf{Generation of BHRF1 binders across 4 epitope-conditioning approaches.} \textbf{(A)} Terminal AF2-M ipTM distributions across baseline, unallocated, and dynamic allocation methods. Dashed lines denote group medians (grey: Vanilla). \textbf{(B)} Budget allocation across targeted epitopes and the unconditional (Hotspot off) baseline ($1{,}728$ backbones per method across $3$ replicates of $576$). \textbf{(C)} Designable backbone yield. Segments detail absolute counts and condition-specific success rates. \textbf{(D)} Compute breakdown (GPU-hours) and efficiency (designable backbones per GPU-hour). \textbf{(E)} Structural visualization of targeted BHRF1 epitopes.}
  \label{fig:fig3}
\end{figure}

\section{Results and Discussion}

\begin{wraptable}{r}{0.57\textwidth}
    \centering
    \setlength{\tabcolsep}{5pt}
    \small
    \caption{\textbf{Compute efficiency and success rate of inference-time steering.} Designable backbones per end-to-end GPU-hour and pass rate (\%) across three targets.}
    \label{tab:Table1}
    \begin{tabular}{@{}lrrr@{}}
\toprule
& & \multicolumn{2}{c}{Steered} \\
\cmidrule(lr){3-4}
Target (Scaffold) & Vanilla & Fixed & ESS \\
\midrule
\multicolumn{4}{@{}l}{\textit{Designable / GPU-h}} \\
PD-L1 (7eow)& 10.6$\pm$0.9 & \textbf{16.9$\pm$0.6} & 14.4$\pm$0.5 \\
IL-7R$\alpha$ (7xl0)& 9.4$\pm$0.3 & \textbf{17.8$\pm$1.2} & 15.3$\pm$1.3 \\
BHRF1 (7eow)& 4.8$\pm$0.3 & \textbf{7.0$\pm$2.8} & 6.3$\pm$0.8 \\
\midrule
\multicolumn{4}{@{}l}{\textit{Pass Rate (\%)}} \\
PD-L1 (7eow)& 14.9$\pm$1.2 & \textbf{42.7$\pm$1.1} & 36.7$\pm$1.2 \\
IL-7R$\alpha$ (7xl0)& 22.9$\pm$0.7 & \textbf{62.2$\pm$4.0} & 53.7$\pm$4.4 \\
BHRF1 (7eow)& 9.7$\pm$0.6 & \textbf{20.2$\pm$8.1} & 18.1$\pm$2.2 \\
\bottomrule
\end{tabular}
     
\end{wraptable}

\subsection{Steering Increases Yield Without Sacrificing Structural Diversity}
Steering on fixed-scaffold nanobody generation was assessed for three distinct targets: PD-L1, BHRF1, and IL-7R$\alpha$. Both fixed-cadence and adaptive (ESS) steering shifted generated candidate distributions toward higher ipTM compared to the unsteered (vanilla) baseline for both terminal ESMFold2 and post-hoc AF2-M oracles (Figure \ref{fig:fig2}).  We also found steering improves the rate of designable backbones yielded per GPU-hour (Table \ref{tab:Table1}). A breakdown of generation and evaluation compute (Figure \ref{fig:appendix:fig2_compute}) and an assessment of the impact of the number of inverse folded sequences evaluated per backbone in identifying passing designs (Figure \ref{fig:appendix:S=n}) are presented in the Appendix.

While steering collapses paired CDR-H3 C$\alpha$ RMSD, a measure of backbone diversity, within individual trajectories, this loss of variation is localized to single runs (Figure \ref{fig:fig2}D). Across independent seeds, steering recovers structural variation by exploring distinct, high-reward conformational basins. Thus, steering does not inherently sacrifice global diversity; rather, it highlights the need to distribute compute across multiple starting conditions to sample the viable paratope space.

However, over-optimizing on an in-loop metric exposes generation to reward hacking. While the terminal ESMFold2 oracle and the post-hoc AF2-M validation oracle generally correlate, this agreement is imperfect (Figure \ref{fig:fig2}A,B,C). Structural overlays of PD-L1 candidates (Figure \ref{fig:fig2}E) serve to highlight instances of agreement and divergence, emphasizing that satisfying the in-loop metric does not guarantee terminal validation success.

\subsection{Dynamic Allocation Maximizes Yield Across Configurations}
To evaluate resource allocation across candidate conditions, each method was evaluated on a fixed computational budget distributed across three independent replicates. For BHRF1, this totaled 1,728 designs per method across four epitope configurations (576 per replicate; Figure \ref{fig:fig3}; Table \ref{tab:bhrf1-epitopes}); for PD-L1, this totaled 2,880 designs per method across six scaffolds (960 per replicate; Figure \ref{fig:fig4}; Table \ref{tab:scaffolds_combined}). PDB identifiers for scaffolds refer to the source complex structures from which the nanobody frameworks were extracted. Initialized with a uniform starting budget across all conditions, both the SMC and SH allocators successfully shifted terminal ipTM higher. Additionally, adaptive global SMC steering achieved nearly a 5$\times$ increase in designable backbones per GPU-hour over the vanilla baseline (rising from 1.41 to 6.82 for BHRF1, and 2.21 to 10.81 for PD-L1). Decoupled SH steering similarly yielded a 2.4–3.1$\times$ throughput improvement. The increase in yield did not sacrifice global conformational diversity across independent runs (Appendix Figures \ref{fig:appendix:fig3_diversity}, \ref{fig:appendix:fig4_diversity}). 

To isolate the distinct contributions of trajectory steering and condition allocation, we evaluated several algorithmic ablations (Appendix Figures \ref{fig:appendix:fig3_ablations} and \ref{fig:appendix:fig4_ablations}). Applying the SH allocation framework to unsteered trajectories improved yield over the vanilla baseline, indicating that iterative condition selection alone provides benefits. However, unsteered SH yielded fewer passing designs than steered SH, demonstrating that the budget allocation and inference-time steering are additive. Furthermore, we evaluated a fixed-cadence SMC allocation policy. While the fixed-cadence approach successfully outperformed both the vanilla and steered-unallocated baselines, the adaptive SMC policy (triggered dynamically by ESS weight degeneracy) produced the highest throughput overall.

\subsection{Oracle Divergence and the Hierarchy of Reward Hacking}
\begin{figure}[!ht]
  \centering
  \includegraphics[width=.7\textwidth]{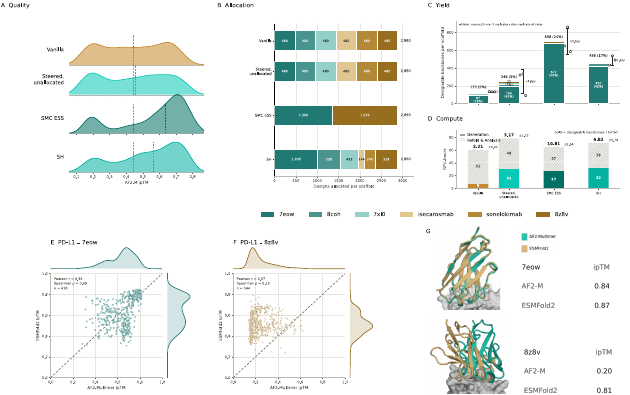}
  \caption{\textbf{Dynamic scaffold allocation and the impact of oracle divergence.} Generation of PD-L1 binders across six distinct scaffolds. \textbf{(A)} Terminal AF2-M ipTM distributions across baseline, unallocated, and dynamic allocation methods. \textbf{(B)} Budget allocation per scaffold ($2,880$ backbones per method across $3$ replicates of $960$ each). \textbf{(C)} Designable backbone yield, detailing absolute counts and scaffold-specific success rates. \textbf{(D)} Compute breakdown (GPU-hours) and efficiency (designable backbones per GPU-hour). \textbf{(E, F)} Oracle correlation (terminal ESMFold2 vs. AF2-M ipTM) for 7eow and 8z8v. \textbf{(G)} Structural overlays showing oracle agreement (7eow) and divergence (8z8v).}
  \label{fig:fig4}
\end{figure}

While joint SMC allocation maximizes raw yield in the PD-L1 binder design experiment, our use of orthogonal oracles for steering and evaluation exposes a vulnerability to reward hacking. SMC allocated 53\% of its particles to the 8z8v-derived framework because of high ESMFold2 ipTM, but yielded zero terminal AF2-M passes due to pose divergence (Figure \ref{fig:fig4}). Decoupled SH mitigates this issue by using orthogonal terminal validation scores for rank cuts. Therefore, SH is preferable when serial rounds are feasible, whereas SMC is suited for settings where in-loop rewards are reliable or offline validation is costly. Importantly, both metrics remain \textit{in silico} surrogates; high ipTM scores do not guarantee \textit{in vitro} affinity ($K_D$) or stability, leaving optimization vulnerable to model-specific artifacts.

\section{Conclusion}
Dynamic resource allocation paired with inference-time steering converts static scaffold and epitope constraints into active optimization variables, increasing \textit{in silico} nanobody yield without sacrificing structural diversity. While global SMC maximizes throughput, decoupled multi-armed bandits offer robustness against in-loop reward hacking. Because intermediate confidence metrics remain surrogates, adapting these allocators to wet-lab-correlated oracles represents a primary direction for future work.

\clearpage 

\section{Acknowledgements}
We would like to thank Mike Schmidt and Piotr Bobrowicz for their scientific support and critical review of the mansucript. We also thank Esteban Gea for his assistance with the computational infrastructure used in early experiments. 

\bibliographystyle{unsrtnat}
\bibliography{refs}

@InProceedings{singhal2025fksteering,
  title     = {A General Framework for Inference-time Scaling and Steering of Diffusion Models},
  author    = {Singhal, Raghav and Horvitz, Zachary and Teehan, Ryan and Ren, Mengye and Yu, Zhou and McKeown, Kathleen and Ranganath, Rajesh},
  booktitle = {Proceedings of the 42nd International Conference on Machine Learning},
  series    = {Proceedings of Machine Learning Research},
  volume    = {267},
  pages     = {55810--55827},
  year      = {2025},
  publisher = {PMLR},
  eprint    = {2501.06848},
  archivePrefix = {arXiv},
  url       = {https://proceedings.mlr.press/v267/singhal25b.html}
}

@article{hartman2025fkprotein,
  title   = {Controllable Protein Design with Particle-Based {Feynman--Kac} Steering},
  author  = {Hartman, Erik and Wallin, Jonas and Malmstr{\"o}m, Johan and Olsson, Jimmy},
  journal = {arXiv preprint arXiv:2511.09216},
  year    = {2025},
  eprint  = {2511.09216},
  archivePrefix = {arXiv},
  doi     = {10.48550/arXiv.2511.09216}
}

@article{stark2025boltzgen,
  title   = {{BoltzGen}: Toward Universal Binder Design},
  author  = {Stark, Hannes and Faltings, Felix and Choi, MinGyu and Xie, Yuxin and Hur, Eunsu and O'Donnell, Timothy and Bushuiev, Anton and U{\c{c}}ar, Talip and Passaro, Saro and Mao, Weian and Reveiz, Mateo and Bushuiev, Roman and Pluskal, Tom{\'a}{\v{s}} and Sivic, Josef and Kreis, Karsten and Vahdat, Arash and Wohlwend, Jeremy and Corso, Gabriele and Barzilay, Regina and Jaakkola, Tommi},
  journal = {bioRxiv},
  year    = {2025},
  note    = {Posted November 2025},
  doi     = {10.1101/2025.11.20.689494}
}

@article{zambaldi2024alphaproteo,
  title   = {De novo design of high-affinity protein binders with {AlphaProteo}},
  author  = {Zambaldi, Vinicius and La, David and Chu, Alexander E. and Patani, Harshnira and Danson, Amy E. and Kwan, Tristan O. C. and Frerix, Thomas and Schneider, Rosalia G. and Saxton, David and Thillaisundaram, Ashok and Wu, Zachary and Moraes, Isabel and Lange, Oskar and Papa, Eliseo and Hassabis, Demis and Kohli, Pushmeet and Fergus, Rob and Wang, Jue},
  journal = {arXiv preprint arXiv:2409.08022},
  year    = {2024},
  eprint  = {2409.08022},
  archivePrefix = {arXiv},
  doi     = {10.48550/arXiv.2409.08022}
}

@article{vincke2009general,
  title   = {General strategy to humanize a camelid single-domain antibody and identification of a universal humanized nanobody scaffold},
  author  = {Vincke, C{\'e}cile and Loris, Remy and Saerens, Dirk and Martinez-Rodriguez, Sergio and Muyldermans, Serge and Conrath, Katja},
  journal = {Journal of Biological Chemistry},
  volume  = {284},
  number  = {5},
  pages   = {3273--3284},
  year    = {2009},
  doi     = {10.1074/jbc.M806889200}
}

@article{rossotti2022immunogenicity,
  title   = {Immunogenicity and humanization of single-domain antibodies},
  author  = {Rossotti, Martin A. and B{\'e}langer, Kasandra and Henry, Kevin A. and Tanha, Jamshid},
  journal = {The FEBS Journal},
  volume  = {289},
  number  = {14},
  pages   = {4304--4327},
  year    = {2022},
  doi     = {10.1111/febs.15809}
}

@article{bennett2026atomically,
  title = {Atomically accurate de novo design of antibodies with {RFdiffusion}},
  author = {Bennett, Nathaniel R. and Watson, Joseph L. and Ragotte, Robert J. and Borst, Andrew J. and See, DeJenae L. and Weidle, Connor and Biswas, Riti and Yu, Yutong and Shrock, Ellen L. and Ault, Russell and Leung, Philip J. Y. and Huang, Buwei and Goreshnik, Inna and Tam, John and Carr, Kenneth D. and Singer, Benedikt and Criswell, Cameron and Wicky, Basile I. M. and Vafeados, Dionne and Garcia Sanchez, Mariana and Kim, Ho Min and Vazquez Torres, Susana and Chan, Sidney and Sun, Shirley M. and Spear, Timothy},
  journal = {Nature},
  volume = {649},
  number = {8095},
  pages = {183--193},
  year = {2026},
  note = {Published online 5 November 2025},
  doi = {10.1038/s41586-025-09721-5}
}

@article{millefragoso2026germinal,
  title   = {Efficient generation of epitope-targeted antibodies with {Germinal}},
  author  = {Mille-Fragoso, Luis S. and Driscoll, Claudia L. and Wang, John N. and Dai, Haoyu and Widatalla, Talal and Zhang, Jim L. and Zhang, Xiaowei and Rao, Bing and Feng, Liang and Hie, Brian L. and Gao, Xiaojing J.},
  journal = {Nature Biotechnology},
  year    = {2026},
  doi     = {10.1038/s41587-026-03187-0}
}

@article{skreta2025feynman,
  title={Feynman-kac correctors in diffusion: Annealing, guidance, and product of experts},
  author={Skreta, Marta and Akhound-Sadegh, Tara and Ohanesian, Viktor and Bondesan, Roberto and Aspuru-Guzik, Al{\'a}n and Doucet, Arnaud and Brekelmans, Rob and Tong, Alexander and Neklyudov, Kirill},
  journal={arXiv preprint arXiv:2503.02819},
  year={2025}
}

@inproceedings{yoo2026searchbinder,
  title     = {Search-Based Inference-Time Scaling for All-Atom Protein Binder Design},
  author    = {Yoo, Kiwoong and Yim, Soorin and Lee, Kyungwook and Park, Sungjoon and Kong, Jaehyeong and Hwang, Doyeong and Jang, Jongseong and Kim, Kiyoung},
  booktitle = {ICLR 2026 Workshop on Generative and Experimental Perspectives for Biomolecular Design (GEM)},
  year      = {2026},
  url       = {https://openreview.net/forum?id=qLWzokU9E7},
  note      = {Concurrent work}
}

@article{efron2011tweedie,
  title   = {Tweedie's Formula and Selection Bias},
  author  = {Efron, Bradley},
  journal = {Journal of the American Statistical Association},
  volume  = {106},
  number  = {496},
  pages   = {1602--1614},
  year    = {2011},
  doi     = {10.1198/jasa.2011.tm11181}
}

@article{gao2019batched,
  title={Batched multi-armed bandits problem},
  author={Gao, Zijun and Han, Yanjun and Ren, Zhimei and Zhou, Zhengqing},
  journal={Advances in Neural Information Processing Systems},
  volume={32},
  year={2019}
}

@InProceedings{karnin2013successivehalving,
  title     = {Almost Optimal Exploration in Multi-Armed Bandits},
  author    = {Karnin, Zohar and Koren, Tomer and Somekh, Oren},
  booktitle = {Proceedings of the 30th International Conference on Machine Learning},
  series    = {Proceedings of Machine Learning Research},
  volume    = {28},
  pages     = {1238--1246},
  year      = {2013},
  publisher = {PMLR}
}

@article{li2018hyperband,
  title   = {Hyperband: A Novel Bandit-Based Approach to Hyperparameter Optimization},
  author  = {Li, Lisha and Jamieson, Kevin and DeSalvo, Giulia and Rostamizadeh, Afshin and Talwalkar, Ameet},
  journal = {Journal of Machine Learning Research},
  volume  = {18},
  number  = {185},
  pages   = {1--52},
  year    = {2018}
}

@misc{rectorbrooks2026generalmultimodalproteindesign,
      title={General Multimodal Protein Design Enables DNA-Encoding of Chemistry}, 
      author={Jarrid Rector-Brooks and Théophile Lambert and Marta Skreta and Daniel Roth and Yueming Long and Zi-Qi Li and Xi Zhang and Miruna Cretu and Francesca-Zhoufan Li and Tanvi Ganapathy and Emily Jin and Avishek Joey Bose and Jason Yang and Kirill Neklyudov and Yoshua Bengio and Alexander Tong and Frances H. Arnold and Cheng-Hao Liu},
      year={2026},
      eprint={2604.05181},
      archivePrefix={arXiv},
      primaryClass={cs.LG},
      url={https://arxiv.org/abs/2604.05181}, 
}

@inproceedings{
didi2026scaling,
title={Scaling Atomistic Protein Binder Design with Generative Pretraining and Test-Time Compute},
author={Kieran Didi and Zuobai Zhang and Guoqing Zhou and Danny Reidenbach and Zhonglin Cao and Sooyoung Cha and Tomas Geffner and Christian Dallago and Jian Tang and Michael M. Bronstein and Martin Steinegger and Emine Kucukbenli and Arash Vahdat and Karsten Kreis},
booktitle={The Fourteenth International Conference on Learning Representations},
year={2026},
url={https://openreview.net/forum?id=qmCpJtFZra}
}

@article{evans2021protein,
  title={Protein complex prediction with AlphaFold-Multimer},
  author={Evans, Richard and O’neill, Michael and Pritzel, Alexander and Antropova, Natasha and Senior, Andrew and Green, Tim and {\v{Z}}{\'\i}dek, Augustin and Bates, Russ and Blackwell, Sam and Yim, Jason and others},
  journal={biorxiv},
  pages={2021--10},
  year={2021},
  publisher={Cold Spring Harbor Laboratory}
}

@article{dauparas2022robust,
  title={Robust deep learning--based protein sequence design using ProteinMPNN},
  author={Dauparas, Justas and Anishchenko, Ivan and Bennett, Nathaniel and Bai, Hua and Ragotte, Robert J and Milles, Lukas F and Wicky, Basile IM and Courbet, Alexis and de Haas, Rob J and Bethel, Neville and others},
  journal={Science},
  volume={378},
  number={6615},
  pages={49--56},
  year={2022},
  publisher={American Association for the Advancement of Science}
}

@article{bennett2023improving,
  title={Improving de novo protein binder design with deep learning},
  author={Bennett, Nathaniel R and Coventry, Brian and Goreshnik, Inna and Huang, Buwei and Allen, Aza and Vafeados, Dionne and Peng, Ying Po and Dauparas, Justas and Baek, Minkyung and Stewart, Lance and others},
  journal={Nature Communications},
  volume={14},
  number={1},
  pages={2625},
  year={2023},
  publisher={Nature Publishing Group UK London}
}

@article{candido2026language,
  title={Language modeling materializes a world model of protein biology},
  author={Candido, Salvatore and Hayes, Thomas and Derry, Alexander and Rao, Roshan and Lin, Zeming and Verkuil, Robert and Wu, Bryan Z and Lee, Jin Sub and Bruguera, Elise S and Keval, Jehan A and others},
  journal={bioRxiv},
  pages={2026--06},
  year={2026},
  publisher={Cold Spring Harbor Laboratory}
}

\clearpage

\appendix

\section{Extended Methods and Implementation Details}

\subsection{Theoretical Considerations and Steering Implementation}
\label{appendix:steering_background}

\paragraph{Base Measure}
As noted in the main text, Feynman-Kac steering reweights base samples toward $p_{\text{target}}(x_{0:T}) \propto q_\theta(x_{0:T})\exp(\Phi(x_{0:T})/\tau)$, where the path functional $\Phi(x_{0:T})$ accumulates per-event \texttt{immediate} potentials $R_t$ along particle genealogies. This potential tilt acts at discrete intermediate resampling events rather than solely at terminal generation. Specific to BoltzGen, the base measure $q_\theta$ reflects the deployed design sampler. Because this sampler is sharpened and under-noised relative to the exact reverse SDE assumed by standard Feynman-Kac guarantees, steering converges to the reward-tilted deployed sampling distribution. Consequently, a given temperature $\tau$ acts more aggressively than in an exact continuous-time process.

\paragraph{Network Time}
Guidance events are placed on a schedule uniform in network time $c(\sigma) = \frac{1}{4}\log\left(\frac{\sigma}{\sigma_{\text{data}}}\right)$ rather than raw diffusion steps, ensuring the schedule remains invariant to the overall step budget. Because $c(\sigma)$ non-linearly stretches the noise schedule, guidance events cluster toward the lower-noise stages of the trajectory.

\paragraph{Steering Parameters}
Steering parameters were empirically optimized for each experiment. For standard unallocated steered runs and steered SH runs, we utilize a population of $N=16$ particles, a value chosen empirically following a sweep of 8, 16, 32, and 64 particles. In joint allocation experiments, the total particle count is scaled to $K \times 16$, where $K$ represents the total number of candidate conditions (scaffolds or epitopes). The steering temperature, $\tau \in [0.1, 0.25]$, was determined via a grid search. Between 6 and 12 guidance events were applied per trajectory, with the exact number tailored to the specific experimental setup. Furthermore, based on an initial parameter sweep ranging from 40\% to 60\%, steering was consistently initiated at the 60\% mark of the diffusion trajectory. We opt for systematic resampling over multinomial. 

Finally, for all experiments, per-event reward increments are computed using the \texttt{immediate} potential mode \citep{hartman2025fkprotein}, where the instantaneous Tweedie estimate evaluates the current reward. Following an initial sweep across \texttt{immediate}, \texttt{difference}, \texttt{sum}, and \texttt{max} potentials~\citep{singhal2025fksteering}, the \texttt{immediate} mode was selected due to higher empirical yield. However, because instantaneous \texttt{immediate} potentials do not construct a standard telescoping Feynman-Kac path functional, standard convergence guarantees do not apply. The realized target distribution instead depends on the algorithmic context: for independent steering and Successive Halving (SH) baselines, the weights are memoryless and reflect only the instantaneous immediate potential, whereas the Global SMC allocator recursively accumulates these immediate increments to target an additive path functional.

\subsection{Global SMC Scaffold Allocation}
\label{app:mixture}

To address the configuration bottleneck, we treat the scaffold or epitope label $k$ as a coordinate of the particle state $z = (k, x_t)$. Let $q_k(x_{0:T})$ be scaffold $k$'s frozen generative law and $\boldsymbol{\omega}$ be a prior over scaffolds. A joint population of $N$ particles targets the reward-tilted joint distribution:
\begin{equation}
\pi(k, x_{0:T}) \propto \omega_k \, q_k(x_{0:T}) \exp(\Phi(x_{0:T})/\tau)
\end{equation}
where $\Phi(x_{0:T}) = \sum_m R(\hat{x}_{0,m})$ is the additive reward path functional. Although the per-event increments are evaluated using the memoryless \texttt{immediate} potential mode, the weight recursion of the Global SMC algorithm accumulates these instantaneous rewards, yielding a target distribution proportional to their cumulative sum. By formulating the label as a component of the state, we deploy standard Sequential Monte Carlo (SMC) on the augmented space. Unlike independent steering, the weight recursion normalizes over a global denominator coupling all $N$ particles across all $K$ candidate conditions (Equation \ref{eq:logw_update}).

\paragraph{Emergent Budget Allocation.}
This global normalization serves as the primary allocation mechanism. The component mass $\rho_k = \sum_{i: k^i=k} w^i_m$ converges to the label marginal of the target joint distribution. No explicit computational budget is assigned. During global systematic resampling, the per-scaffold budget emerges dynamically as the integer count $b_k \approx N\rho_k$. Scaffolds with consistently low in-loop rewards transition to an absorbing state of $b_k = 0$, while highly designable scaffolds duplicate.

\begin{algorithm}[!ht]
\caption{Global SMC scaffold allocator.}
\label{alg:mixture}
\begin{algorithmic}[1]
\Require scaffolds $\{1,\dots,K\}$ with initial counts $b_k$, $\textstyle\sum_k b_k = N$; steps $T$;
potential mode setting the exponent $g$; fixed temperature $\tau$; scaffold
prior $\boldsymbol{\omega}$; warmup events $W$; stop fraction $\eta$; draw rule \textsc{Draw}
\State $\log w^i \gets \log \omega_{k^i} - \log b_{k^i}$, then normalize; \quad $m \gets 0$;
\For{$t = 1$ to $T$}
  \State advance every active particle by one denoising step with local resampling disabled
  \If{$t$ is a shared reward event}
    \State $m \gets m + 1$ \Comment{weight-event clock}
    \State $g^i \gets$ potential-mode exponent of particle $i$ \Comment{$g^i\!=\!R(\hat{x}_{0,m}^i)$ for \texttt{immediate}}
    \State $\Delta \gets \log \sum_j \exp \big(\log w^j + g^j/\tau\big)$;
    \State $\log w^i \gets \big(\log w^i + g^i/\tau\big) - \Delta$
    \Comment{persistence and global normalization}
    \If{$\textsc{Draw}\big(m,\{w^i\}\big)$ \textbf{and} $m > W$ \textbf{and} $t/T \le \eta$ \textbf{and} $t < T$}
      \State draw $N$ ancestors by one global systematic resample over $\{w^i\}$, particles ordered by scaffold label
      \State regroup ancestors by scaffold label; $\log w^i \gets -\log N$
      \Comment{mass $\to$ multiplicity}
    \EndIf
  \EndIf
\EndFor
\State \Return joint population, emergent counts $b_k=|\{i: k^i=k\}|$, and terminal weights $\{w^i\}$
\end{algorithmic}
\end{algorithm}

\clearpage

\subsection{Successive Halving (SH) Schedule}
\label{app:bracket}
The decoupled allocation approach adapts the Batched Successive Elimination (BaSE) framework \citep{gao2019batched,karnin2013successivehalving}. Allocation is implemented across multiple rounds, where available configurations (arms) are evaluated by steering a small batch of particles. We implement this as a specific case of BaSE with three structural modifications. 

First, rather than utilizing a confidence-bound threshold for elimination, our algorithm removes the confidence width ($\delta = 0$). This reduces the elimination step to an unconditional rank-based cut that prunes the bottom 50\% of configurations at the end of each round. Second, configurations are ranked using the 90th percentile of their terminal validation scores ($q=0.90$) rather than the mean. Finally, while theoretical bounds favor a geometric or minimax budget schedule, we distribute the computational budget uniformly across rounds, as our experiments indicated this can provide greater robustness to the varying quality and composition of the initial panel.

\begin{algorithm}[!ht]
\caption{\textbf{Outer-loop successive-halving allocator over generative configurations.}
Arms are ranked by terminal validation scores of completed designs. Importance weights are never compared across configurations. An equal-budget schedule ($B_m = B_{\text{total}}/M$) is used and $\delta = 0$; a positive elimination width recovers standard batched successive elimination \citep{gao2019batched}.}
\label{alg:bracket}
\begin{algorithmic}[1]
\Require arms $\mathcal{A}$ (configurations); rounds $M$; survival fraction $\kappa$;
round budgets $B_1,\dots,B_M$ with $\sum_m B_m = B_{\text{total}}$ designs; quantile $q$; elimination width $\delta$
\State $\mathcal{S}_k \gets \emptyset$ for every arm $k$ \Comment{running score pool}
\For{$m = 1$ to $M$}
  \ForAll{$k \in \mathcal{A}$} \Comment{split $B_m$ evenly over surviving arms}
    \State generate $B_m / |\mathcal{A}|$ fresh designs from arm $k$ via the memoryless, independent FK sampler
    \State score each with the terminal validation oracle and add the scores to $\mathcal{S}_k$
  \EndFor
  \If{$m < M$} \Comment{eliminate}
    \State $s_k \gets \mathrm{Quantile}_q(\mathcal{S}_k)$ \Comment{over the current round's pooled designs}
    \State sort $\mathcal{A}$ by $s_k$, best first
    \State drop the arms ranked below $\lceil \kappa|\mathcal{A}|\rceil$ whose gap $\max_j s_j - s_k$ is at least $\delta$
  \EndIf
\EndFor
\State \Return every generated design and the surviving arm(s)
\end{algorithmic}
\end{algorithm}

\clearpage

\subsection{Post-Hoc Designability Validation}
\label{appendix:designability}

Generated nanobody candidates are evaluated using a post-hoc structural validation pipeline adapted from the Proteina Complexa framework~\citep{didi2026scaling}, which builds upon the AlphaProteo designability criteria \citep{zambaldi2024alphaproteo}.

For each generated nanobody backbone, we sample eight sequences using ProteinMPNN \citep{dauparas2022robust}, restricting sequence redesign exclusively to the design residues (i.e., the CDRs) while maintaining the fixed scaffold sequence. These redesigned sequences are folded in complex with the target using the ColabDesign implementation of AlphaFold2-Multimer (AF2-M).  Figure \ref{fig:appendix:S=n} details the impact of the number of inverse folding sequences sampled. Following previous structural verification protocols \citep{bennett2023improving}, we enable target templating and utilize the initial guess feature, which initializes the AF2 pair representation using the pairwise distances extracted from the generative model's 3D backbone.

To determine designability, we evaluate each folded structure across three metrics: inter-chain predicted alignment error (ipAE), complex predicted local distance difference test (complex pLDDT), and the self-consistent root-mean-square deviation of the binder (binder scRMSD). A generated backbone is classified as designable if any sequence among its eight ProteinMPNN variants satisfies the following AlphaProteo thresholds:

\begin{enumerate}
    \item ipAE < 7 \AA
    \item complex pLDDT > 90
    \item binder scRMSD < 1.5 \AA
\end{enumerate}

\clearpage

\begin{table}[htbp]
\centering
\begin{threeparttable}
\caption{\textbf{Epitope-conditioning arms for the BHRF1 (PDB \texttt{2WH6}) experiment.}}
\label{tab:bhrf1-epitopes}
\small
\begin{tabular}{@{}l l@{}}
\toprule
Epitope & Nominated hotspot residues (PDB numbering)\\
\midrule
Ep1~\citep{zambaldi2024alphaproteo}         & Phe65, Thr74, Glu77, Asp82, Ser85, Arg93 \\
Ep2~\citep{millefragoso2026germinal}         & Val86, Ile90, Arg93 \\
Ep3         & Glu77, His78, Leu81 \\
\addlinespace
Hotspot off & none nominated \\
\bottomrule
\end{tabular}
\end{threeparttable}
\end{table}

\begin{table}[htbp]
\centering
\begin{threeparttable}
\caption{\textbf{Properties and generative parameters of the evaluated nanobody scaffolds.}}
\label{tab:scaffolds_combined}
\small
\begin{tabular}{@{}l c c c@{}}
\toprule
Scaffold / Source\tnote{a} & FR Identity\tnote{b} & CDR-H3 Len. & Insertions\tnote{c} \\
\midrule
\texttt{8coh} & 85.0\,\% & 19 & 3 / 3 / 7 \\
\texttt{sonelokimab} & 90.0\,\% & 17 & 3 / 4 / 4 \\
\texttt{7eow} & 85.0\,\% & 21 & 3 / 3 / 7 \\
\texttt{7xl0} & 87.5\,\% & 15 & 3 / 3 / 6 \\
\midrule
\texttt{8z8v} & 91.2\,\% & 8 & 3 / 3 / 3 \\
\texttt{isecarosmab} & 88.8\,\% & 8 & 3 / 2 / 2 \\
\bottomrule
\end{tabular}
\begin{tablenotes}[flushleft]\footnotesize
\item[a] PDB identifiers denote the source complex structure from which the nanobody framework backbone was extracted. Named entries (e.g., \textit{sonelokimab}, \textit{isecarosmab}) correspond to clinical nanobody frameworks.
\item[b] Per-position identity of FR1+FR2+FR3 to human germline IGHV3-23.
\item[c] Number of residues inserted by BoltzGen for H1 / H2 / H3.
\end{tablenotes}
\end{threeparttable}
\end{table}

\begin{figure}[!ht]
  \centering
  \includegraphics[width=0.6\textwidth]{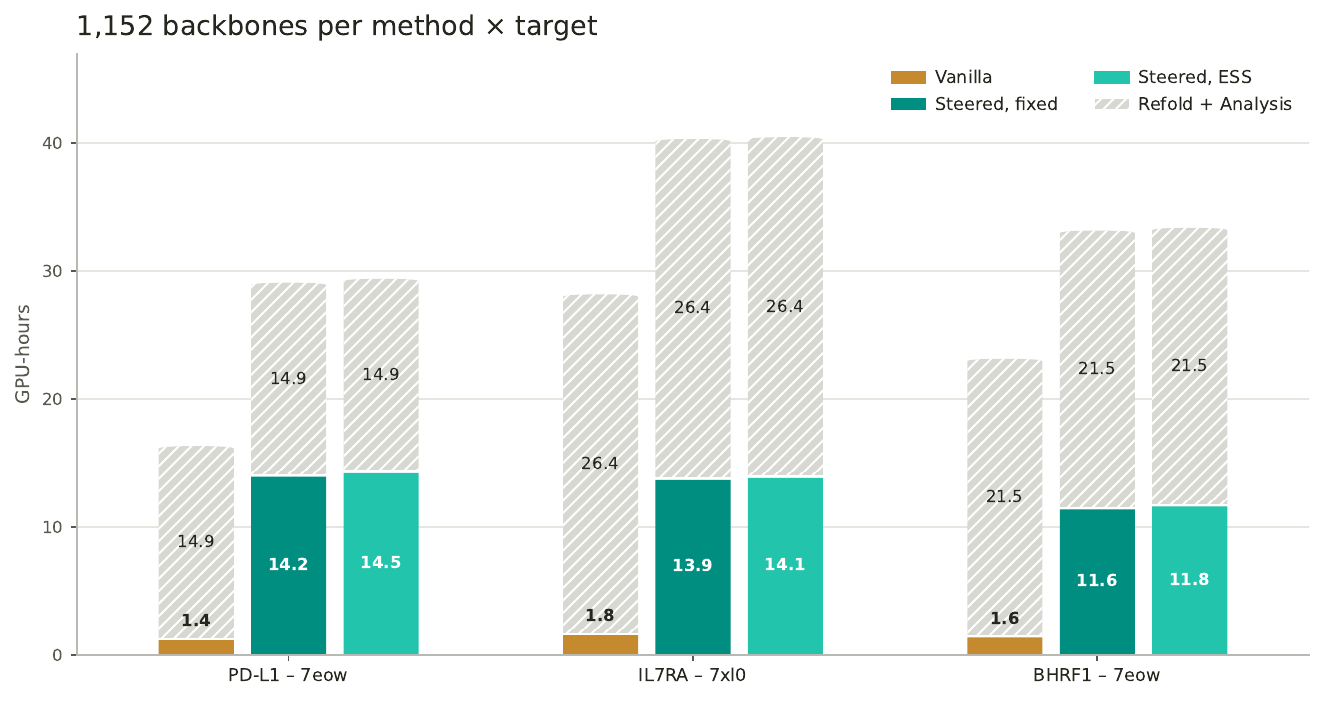}
  \caption{\textbf{Steered and baseline compute overhead breakdown by generation method and target for three targets.} Total GPU-hours expended for 1,152 backbones per target across steering policies. Solid bars represent direct steered/vanilla trajectory generation, while hatched bars account for in-loop and terminal refolding, inverse folding, and post-hoc filtering compute.}
  \label{fig:appendix:fig2_compute}
\end{figure}

\begin{figure}[!ht]
  \centering
  \includegraphics[width=\textwidth]{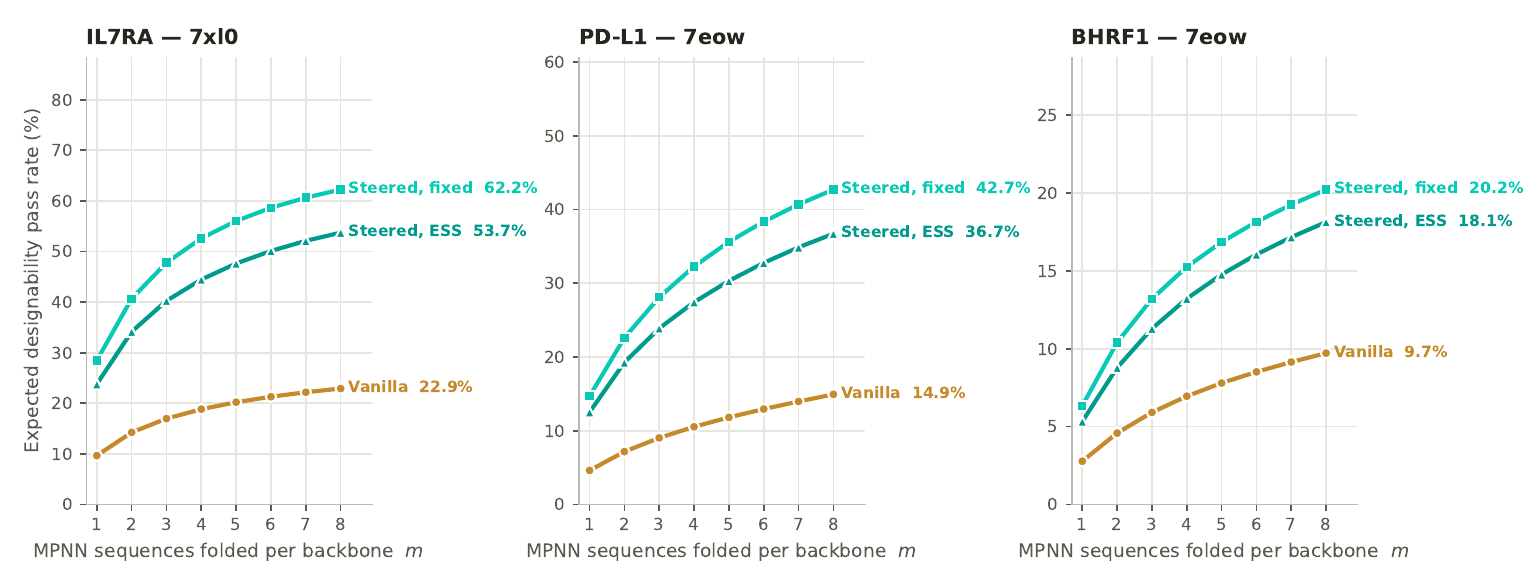}
  \caption{\textbf{Expected designability pass rate vs. MPNN sequence budget (Unallocated).} The expected pass rate (y-axis) is the fraction of backbones yielding $\ge 1$ successful design out of $m$ folded sequences (x-axis, $m=1..8$). 1,152 backbone designs per (target, scaffold, method) tuple.}
  \label{fig:appendix:S=n}
\end{figure}

\begin{figure}[!ht]
  \centering
  \includegraphics[width=\textwidth]{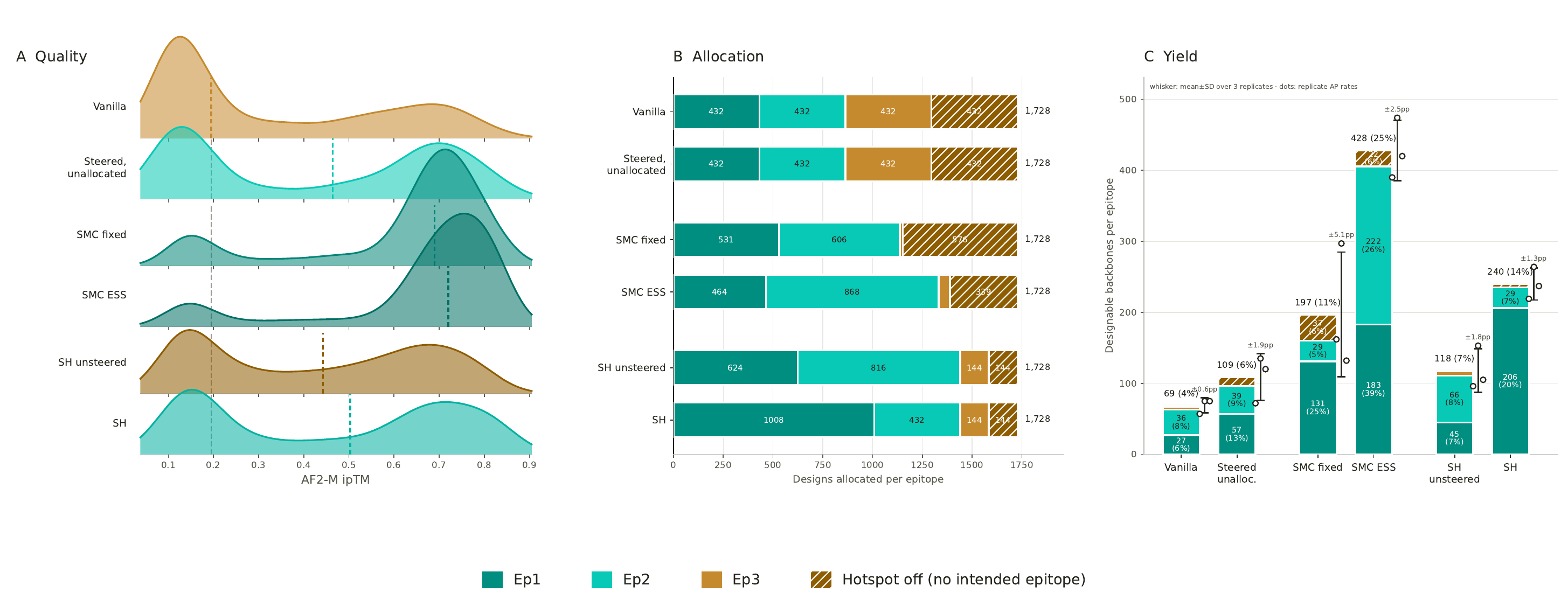}
    \caption{\textbf{Ablation benchmark for BHRF1 binder generation across epitope-conditioning approaches.} \textbf{(A)} Terminal AF2-M ipTM distributions across baselines, dynamic allocation methods, and ablation variants (SMC fixed resampling cadence and SH unsteered). Dashed lines denote group medians (grey: Vanilla). \textbf{(B)} Computational budget allocation across targeted epitopes (Ep1–Ep3) and the unconditional baseline (Hotspot off). Total design budget is held constant at $1,728$ trajectories per method across $3$ replicates with $576$ designs each. \textbf{(C)} Designable backbone yield evaluated under post-hoc structural validation criteria. Bar segments specify absolute counts and condition-specific success rates per epitope.}
  \label{fig:appendix:fig3_ablations}
\end{figure}

\begin{figure}[!ht]
  \centering
  \includegraphics[width=0.6\textwidth]{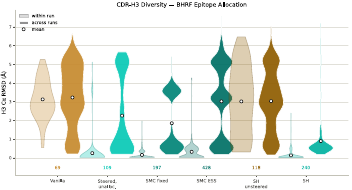}
  \caption{\textbf{CDR-H3 structural diversity for BHRF1 epitope allocation.} C$\alpha$ RMSD of passing designs across baseline, unallocated steering, and dynamic allocation methods (including SMC and SH ablation variants). Lighter violins denote diversity within a single run, whereas darker violins indicate diversity across multiple runs, highlighting that while steering restricts structural diversity within individual trajectories, it successfully explores diverse productive basins across independent runs. Circles denote the mean, and the numbers at the bottom denote the total passing designs per method.}
  \label{fig:appendix:fig3_diversity}
\end{figure}

\begin{figure}[!ht]
  \centering
  \includegraphics[width=\textwidth]{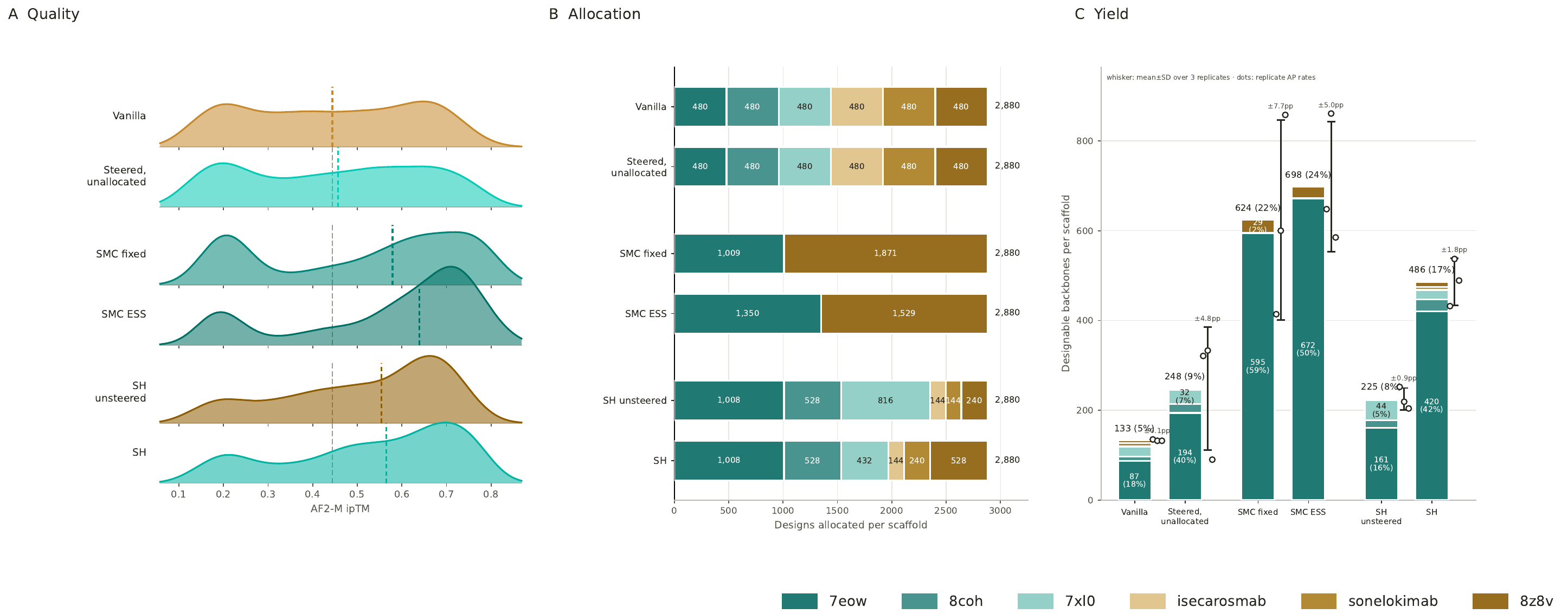}
  \caption{\textbf{Ablation benchmark for PD-L1 binder generation and scaffold allocation.} Evaluation across six candidate scaffolds (extracted from complex structures 7eow, 8coh, 7xl0, 8z8v, and clinical-stage frameworks isecarosmab and sonelokimab). \textbf{(A)} Terminal AF2-M ipTM distributions across baselines, dynamic allocation methods, and ablation variants (SMC global fixed cadence and SH unsteered). Dashed lines denote group medians (grey: Vanilla). \textbf{(B)} Computational budget allocation per scaffold, holding total trajectory count constant at $2,880$ designs per method across $3$ replicates of $960$ designs each. \textbf{(C)} Designable backbone yield under post-hoc structural validation criteria. Bar segments specify absolute counts and scaffold-specific success rates.}
  \label{fig:appendix:fig4_ablations}
\end{figure}

\begin{figure}[!ht]
  \centering
  \includegraphics[width=0.55\textwidth]{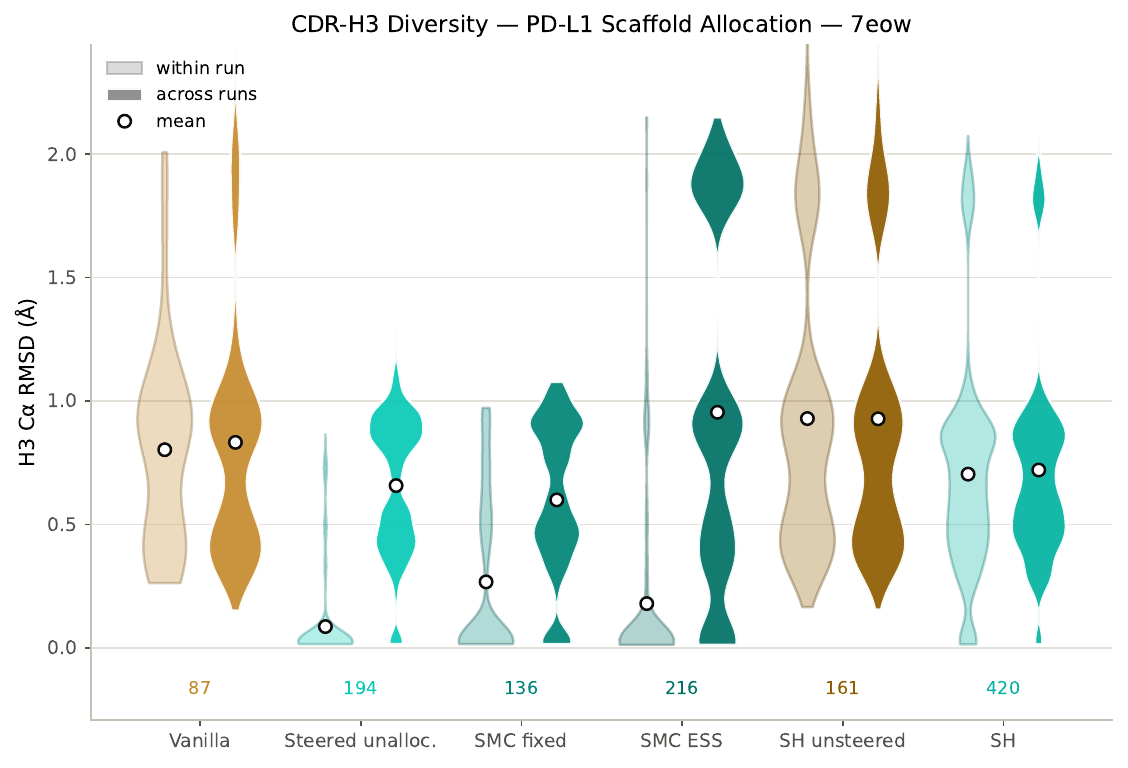}
  \caption{\textbf{CDR-H3 structural diversity for PD-L1 Scaffold allocation.} C$\alpha$ RMSD of passing designs for the 7eow scaffold across baseline, unallocated steering, and dynamic allocation methods (including SMC and SH ablation variants). Lighter violins denote diversity within a single run, whereas darker violins indicate diversity across multiple runs, highlighting that while steering restricts structural diversity within individual trajectories, it successfully explores diverse productive basins across independent runs. Circles denote the mean, and the numbers at the bottom denote the total passing designs per method.}
  \label{fig:appendix:fig4_diversity}
\end{figure}

\end{document}